\documentclass[lettersize,journal]{IEEEtran}
\usepackage{amsmath,amsfonts}
\usepackage{algorithmic}
\usepackage{array}
\usepackage[caption=false,font=normalsize,labelfont=sf,textfont=sf]{subfig}
\usepackage{textcomp}
\usepackage{stfloats}
\usepackage{url}
\usepackage{bm}
\usepackage{verbatim}
\usepackage{graphicx}
\usepackage{multirow}
\usepackage{subfloat}
\usepackage{cite}

\def\BibTeX{{\rm B\kern-.05em{\sc i\kern-.025em b}\kern-.08em
    T\kern-.1667em\lower.7ex\hbox{E}\kern-.125emX}}
\usepackage{balance}
\begin{document}
\title{Improving Speech Translation by Cross-modal Multi-grained Contrastive Learning}
\author{Hao Zhang, Nianwen Si, Yaqi Chen, Wenlin Zhang, Xukui Yang, Dan Qu, Wei-Qiang Zhang, \emph{Senior Member, IEEE}
\thanks{Manuscript received 9, August 2022; revised 5 December 2022 and 29 January 2023; accepted 31 January 2023. Date of publication 13 February 2023; date of current version 24 February 2023. This work was supported in part by the National Natural Science Foundation of China under Grants 62171470 and 62276153, and in part by the Central Plains Science and Technology Innovation Leading Talent Foundation. The associate editor coordinating the review of this manuscript and approving it for publication was Dr. Naoyuki Kanda.
(Corresponding authors: Xukui Yang; Dan Qu.)

Hao Zhang, Nianwen Si, Yaqi Chen, Wenlin Zhang, Xukui Yang, and Dan Qu are with the School of the Information Engineering University, Zhengzhou 450000, China. Email: (haozhang012, snw1608, chyaqi163, wenlinzzz, gzyangxk, qudan\_xd)@163.com.

Wei-Qiang Zhang is with the Department of Electronic Engineering, Tsinghua University, Beijing 100084, China. Email: (wqzhang@tsinghua.edu.cn).

Digital Object Identifier 10.1109/TASLP.2023.3244521}}

\markboth{IEEE/ACM TRANSACTIONS ON AUDIO, SPEECH, AND LANGUAGE PROCESSING, ~Vol.~31, ~2023}%
{How to Use the IEEEtran \LaTeX \ Templates}

\maketitle

\begin{abstract}
The end-to-end speech translation (E2E-ST) model has gradually become a mainstream paradigm due to its low latency and less error propagation. However, it is non-trivial to train such a model well due to the task complexity and data scarcity. The speech-and-text modality differences result in the E2E-ST model performance usually inferior to the corresponding machine translation (MT) model. Based on the above observation, existing methods often use sharing mechanisms to carry out \textbf{\emph{implicit knowledge transfer}} by imposing various constraints. However, the final model often performs worse on the MT task than the MT model trained alone, which means that the knowledge transfer ability of this method is also limited. To deal with these problems, we propose the FCCL (\underline{F}ine- and \underline{C}oarse- Granularity \underline{C}ontrastive \underline{L}earning) approach for E2E-ST, which makes \emph{explicit knowledge transfer} through cross-modal multi-grained contrastive learning. A key ingredient of our approach is applying contrastive learning at both sentence- and frame-level to give the comprehensive guide for extracting speech representations containing rich semantic information. In addition, we adopt a simple whitening method to alleviate the representation degeneration in the MT model, which adversely affects contrast learning. Experiments on the MuST-C benchmark show that our proposed approach significantly outperforms the state-of-the-art E2E-ST baselines on all eight language pairs. Further analysis indicates that FCCL can free up its capacity from learning grammatical structure information and force more layers to learn semantic information.
\end{abstract}


\section{Introduction}
\IEEEPARstart{S}{peech} Translation (ST) takes the speech in one language (source) as input and outputs the translated text in another language (target). Traditional ST systems \cite{ney1999speech,mathias2006statistical} cascade automatic speech recognition (ASR) and machine translation (MT), which might suffer from error propagation and high latency. With the rapid progress of deep learning, end-to-end speech translation (E2E-ST) has attracted more attention \cite{berard2018end,sperber2019attention,liu2019end,dong2021listen,liu2020bridging,du2021regularizing,fang2022stemm,ye2021end,xu2021stacked,han2021learning}. Different from the traditional cascading method, which decomposes ST into two sub-tasks, E2E-ST jointly handles them in a single neural network, which endows it with unique advantages, such as less error propagation and fewer parameters.  

\begin{figure}[!t]
\centering
\includegraphics[scale=0.6]{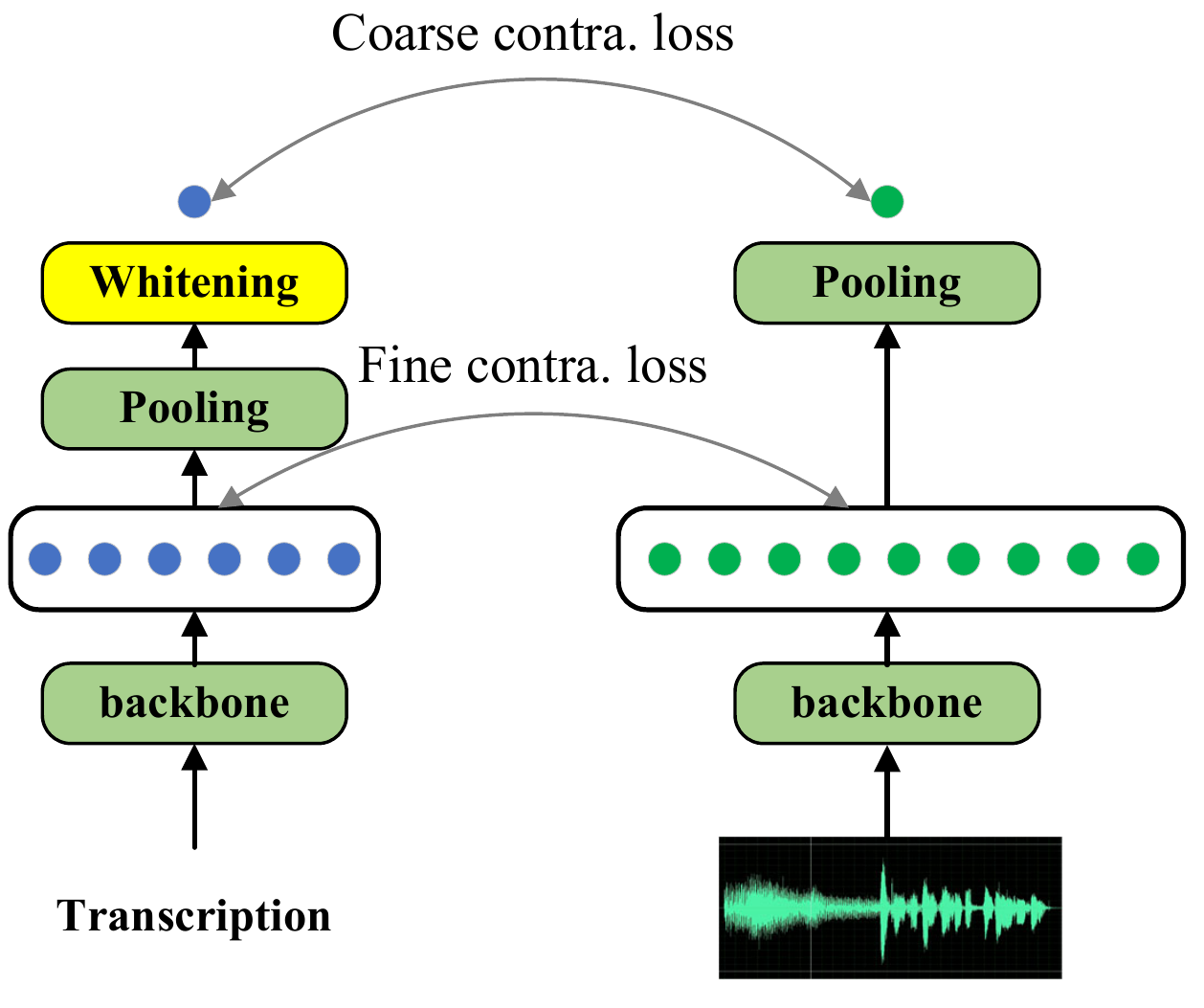}
\caption{Schematic diagram of multi-grained contrastive learning. Text and speech are inputted into the backbone to get the token-level (frame-level) representations. The number of blue and green points before pooling represents the sequence length. We then pool the token-level representations over the time dimension to obtain the sentence-level representations. We conduct fine and coarse granularity contrastive learning at the token- and sentence-level representations, respectively.}
\label{fig:fc}
\end{figure}

However, compared with MT and ASR tasks, E2E-ST is a cross-modal translation task, and its training data is more challenging to collect. The task complexity and data scarcity mean that it is non-trivial to train such a model well. Although both ST\footnote{If not special specified, ST mentioned below refers to E2E-ST by default.} and MT are translation tasks, the performance of the ST model is usually much inferior to the corresponding text-based MT model \cite{liu2020bridging} with the same source and target languages. This can be attributed to the modality gap between speech and text. Compared to discrete text, speech is fine-grained and contains more noise, making it more challenging to extract speech representations containing rich semantic information. Based on the above observation, existing methods focus on transferring knowledge from the MT to the ST model with sophisticated techniques, such as progressive training \cite{ye2021end}, triangular decomposition agreement \cite{du2021regularizing}, and manifold mixup \cite{fang2022stemm}. The common idea behind these methods is to implicitly constrain the parameter space of the ST model by treating the MT task as a constraint term with the sharing mechanism, which can be called \textbf{\emph{implicit knowledge transfer}}. Although impressive performance improvement has been achieved, the performance of the final model in the MT task is often inferior to that of the MT model trained alone \cite{tang2021improving,fang2022stemm}, which means the auxiliary knowledge from the MT task will be less and less as the training progresses. Consequently, implicitly transferring knowledge may not be optimal for the ST task.

Recalling the original problem mentioned above, the modality gap makes it hard to extract semantically rich representation from speech. On the contrary, it is easy to obtain semantic information from text due to its structural advantage. Why not use text representation to provide direct guidance for extracting speech representation? In this paper, we propose FCCL (\underline{F}ine- and \underline{C}oarse- Granularity \underline{C}ontrastive \underline{L}earning) to transfer explicit knowledge from the MT to the ST model. The fundamental motivation behind our method is that since the semantic space is shared between different modalities, the speech and text with similar semantics should be close in the semantic space. In contrast, data pairs with different meanings should be pushed far away. Furthermore, the continuous semantic space of speech should be linked with the discrete symbolic space of text, which is more flexible in transferring knowledge.

Specifically, as shown in Figure \ref{fig:fc}, we provide the comprehensive guidance for extracting speech representations from both frame- and sentence-level to bridge the modality gap. The encoder output of the ST model is regarded as the frame-level speech representation. We average the encoder output along the time dimension to get the sentence-level speech representation. Similar to the ST model, we can get the token- and sentence-level text representation for the MT model. Coarse granularity contrastive learning is conducted at sentence-level. ST is a generative task, which means each speech frame in the encoder should have precise semantics. The representation learned from sentence-level can be sub-optimal for frame-level. Thus, we conduct fine granularity contrastive at frame-level (token-level) to learn more refined representations. 

It is straight to implement coarse granularity contrast. The fine granularity contrast, however, is not the same. To implement fine granularity contrast, we should obtain the alignment between the speech frame and text token. A common way to find this correspondence is to carry out force alignment through an additional alignment model \cite{fang2022stemm}. However, its computational complexity is high, and the alignment model is not always available. To deal with this problem, we propose a maximum similarity method to get the alignment in an unsupervised manner. Based on the obtained alignment, we conduct fine granularity contrastive to find the optimal the token-level representation for the decoder.

Our main contributions are summarized as follows:
\begin{enumerate}
\item We propose FCCL, a cross-modal multi-grained contrastive learning method, to conduct \textbf{\emph{explicit knowledge transfer}} from the MT to the ST model.
\item We propose a maximum similarity method to effectively and efficiently find the correspondence between speech frames and text tokens in an unsupervised manner, which needs negligible computation overhead.
\item We use a whitening method to alleviate the representation degeneration of the MT model by transforming the sentence representations into a standard normal distribution, which satisfies isotropy.
\item We show through CCA analysis that FCCL can free up its capacity from learning grammatical structure information and force more layers to learn semantic information.
\item We conduct experiments on the MuST-C benchmark on all eight language pairs. The experiment results and detailed analysis verify the effectiveness of our proposed method.
\end{enumerate}

\section{Related Works}
\noindent\textbf{End-to-end ST} Early ST system cascaded the ASR and MT system. Benefiting from the development of deep learning in recent years, E2E-ST \cite{berard2016listen,duong2016attentional} has become a mainstream paradigm due to its advantages, such as lower latency, alleviation of error propagation, and fewer parameters. However, due to the scarcity of triplet training data and the complexity of cross-modal translation task, it is non-trivial to train such a model well. To overcome the scarcity of training data, various methods have been explored, such as data augmentation \cite{kocabiyikoglu2018augmenting,jia2019leveraging,pino2019harnessing}, multi-task learning \cite{weiss2017sequence,liu2020synchronous}, sub-module pre-training \cite{bansal2019pre,stoian2020analyzing,wang2020curriculum,zhang2022speechut,tang2022unified}, self-training \cite{pino2020self}, meta-learning \cite{indurthi2019data}, and interactive decoding \cite{liu2020synchronous}. Meanwhile, some works focus on alleviating the task complexity. One branch notices that the encoder of the ST model is overburdened. They decouple the ST encoder into an acoustic encoder and a semantic encoder to improve the ability to extract information from the speech feature \cite{dong2021listen,liu2020bridging,xu2021stacked}. Another branch aims to bridge the modality gap between speech and text. They focus on transferring knowledge from the MT to the ST model with sophisticated techniques, such as progressive training \cite{ye2021end}, triangular decomposition agreement \cite{du2021regularizing} and manifold mixup \cite{fang2022stemm}. In this work, we explore how to bridge the modality gap via \textbf{\emph{explicit knowledge transfer}} based on contrastive learning.

Contrastive learning aims to learn general representations on many unlabeled data. It has extensively promoted progress in computer vision \cite{he2020momentum,chen2020simple}, natural language processing (NLP) \cite{gao2021simcse,yan2021consert,ye2022frequency}, and speech \cite{wang2021multi,zhang2021xlst,xiao2021contrastive}. However, details of view generation are crucial and require careful design \cite{alayrac2020self}.  In contrast, using multiple modalities as different views is simpler and more natural. Some studies extend contrastive learning to the multimodal domain \cite{radford2021learning,wu2022wav2clip,alayrac2020self}. \cite{han2021learning,ye2022cross} introduce a similar idea to design a speech-text cross-modal contrastive learning module in speech translation. However, they suffer from two unique problems.

First, they trained ST and MT tasks together. Unless unique methods are used \cite{tang2021improving}, the joint-trained model often performs worse on the MT task than the MT model trained alone. Additionally, joint training will also limit the use of advanced techniques to improve the MT performance, which is directly related to the text representation quality.

Second, they ignore the characteristics of the ST task. ST is a generative task, which means each speech frame in the encoder should have precise semantics. However, the representation learned from sentence-level can be sub-optimal for frame-level \cite{wang2021dense}. In this work, we propose cross-modal multi-grained contrastive learning to give the comprehensive guide for extracting speech representations containing rich semantic information. We obtain the text representation from a pretrained MT model. This can ensure high-quality text representation during the training.

\noindent\textbf{Fine Granularity Contrastive learning} Contrastive learning is usually conducted on the overall representation of the input. Nevertheless, better overall representation does not guarantee more accurate fine granularity representation \cite{wang2021dense}. It is sub-optimal for generative tasks or dense prediction tasks. In computer vision, some studies perform contrastive learning at pixel-level to learn finer input representations \cite{wang2021dense,zeng2021multi,wang2021fine}. In the ST task, we need to conduct fine granularity contrastive at frame-level so that each speech frame in the encoder has precise semantics. Different from \cite{fang2022stemm}, which gets the correspondence between speech frame and text token by force alignment, we propose a maximum method to obtain the correspondence in an unsupervised manner with negligible latency overhead.
  
\begin{figure}[!t]
\centering
\includegraphics[scale=0.6]{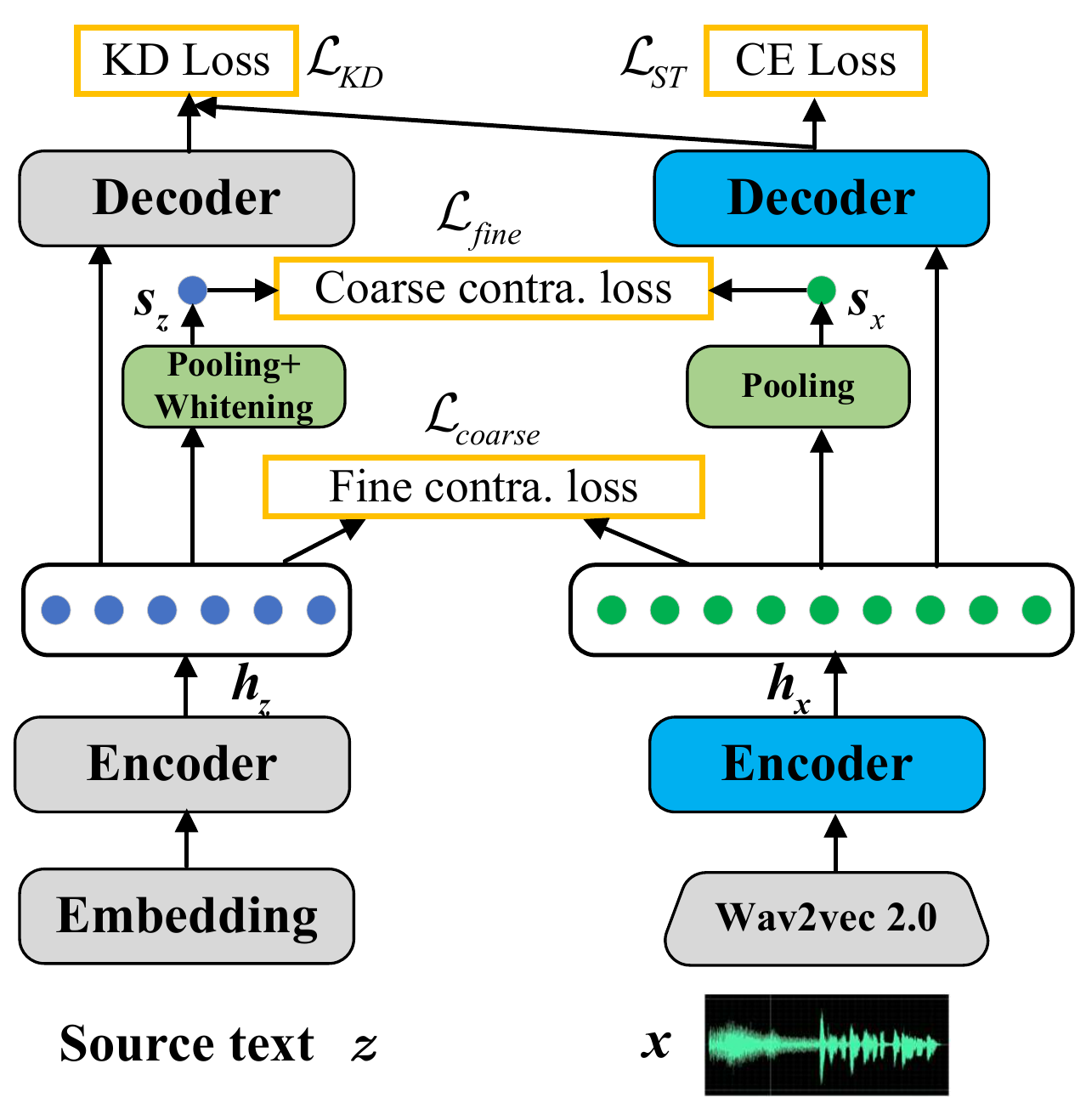}
\caption{Overview of our proposed method. Pooling means averaging the encoder output over the time dimension. FCCL contains three modules, a pretrained MT model, an ST model, and a contrastive module. The grey modules in the figure indicate that its parameters are no longer updated during training. The contra. in the picture is the abbreviation of contrastive. During inference, only the ST model is preserved, and all
other modules are discarded.}
\label{fig:main_work}
\end{figure}

\section{Method}
\noindent In this section, we begin with the fundamental problem formulation of E2E-ST. Then we introduce the coarse and fine granularity contrastive learning in Sections \ref{sec:Coarse Granularity Contrastive Learning} and \ref{sec:Fine Granularity Contrastive Learning}, respectively. The overall structure is shown in Figure \ref{fig:main_work}.

\subsection{Problem Formulation}
The speech translation corpus usually contains \emph{speech-transcription-translation} triples, denoted as ${\cal D} = \{ ({x^{(n)}},{y^{(n)}},{z^{(n)}})\} _{n = 1}^N$, where $x,y,z$ are the audio, the translation in the target language, and the corresponding transcription in the source language, respectively. During training, we optimize the maximum likelihood estimation loss on the training set: 

\begin{equation}
{{{\cal L}}_{ST}}(\theta ) =  - \sum\limits_{(x,y) \in {{\cal D}}}^{} {\log p(y|x)}
\end{equation}

\subsection{Model Architecture}
FCCL contains an MT model, a contrastive learning module, and an ST model. We freeze the parameters of the MT model to avoid performance drop during training. And this manner can also facilitate using sophisticated methods to improve the MT model performance, which can provide better text representations. 

\noindent\textbf{Wav2vec 2.0 as feature extractor} Traditional acoustic features cause model performance degradation when training data is insufficient, since the manually customized process will inevitably lead to information loss \cite{dong2021unist}.  Following the previous works \cite{fang2022stemm,ye2021end,han2021learning}, we adopt Wav2vec 2.0 \cite{baevski2020wav2vec} without finetune to extract speech representation for raw speech $x$. Its parameters are frozen to facilitate quick experiments.  

\subsection{Coarse Granularity Contrastive Learning}
\label{sec:Coarse Granularity Contrastive Learning}
Contrastive learning aims to learn general representations from two views of the same input \cite{he2020momentum,chen2020simple}. However, details of view generation are crucial and require careful design \cite{alayrac2020self}. In contrast, using multiple modalities as different views is simpler and more natural. In this paper, we treat speech and the corresponding transcription as expressions of the same semantics in different modalities.

\begin{figure}[!t] 
  \centering 
  \subfloat[isotropy]{ 
    \label{fig:subfig:a} %% label for first subfigure 
    \includegraphics[width=1.4in]{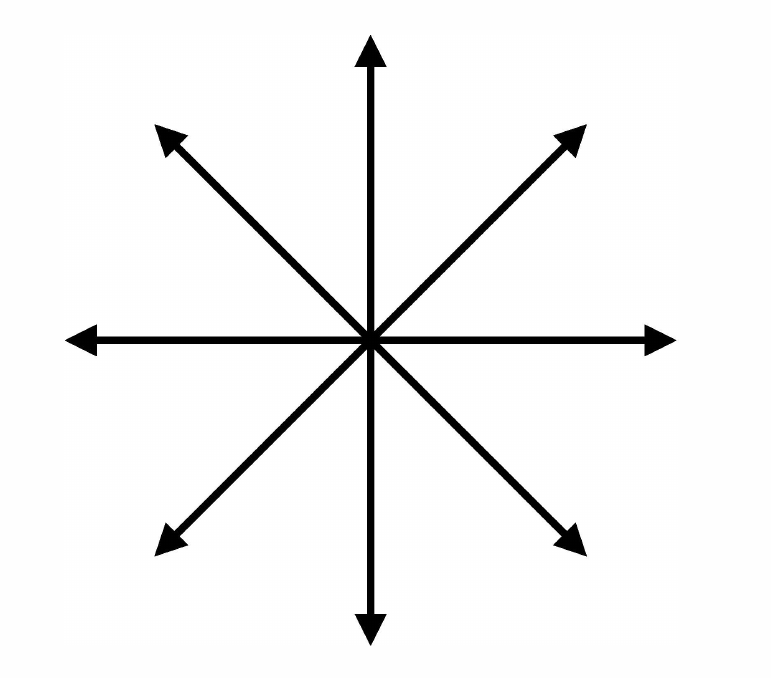}}
  \subfloat[anisotropy]{ 
    \label{fig:subfig:b} %% label for second subfigure 
    \includegraphics[width=1.4in]{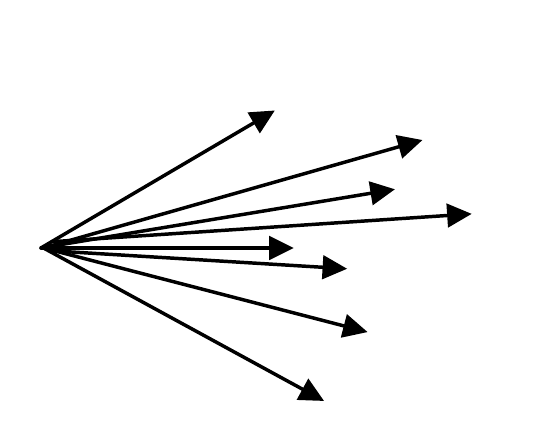}} 
  \caption{Schematic diagram of isotropy and anisotropy. The arrow represents the directions of the word representations.} 
  \label{fig:iso} %% label for entire figure 
\end{figure}

Given the input speech-transcription pairs $\{ ({x^{(n)}},{z^{(n)}})\} _{n = 1}^N$, the encoded representations are $\{ (h_x^{(n)},h_z^{(n)})\} _{n = 1}^N$. We average them in terms of the time dimension to get the sentence-level representation. The natural idea is to compute contrastive loss on the obtained sentence-level representation. However, the representation degeneration \cite{li2020sentence,gao2021simcse} of the MT model makes it problematic. As shown in Figure \ref{fig:iso}a, affected by word frequency, the representations finally learned by the MT model are squeezed into a cone and are not uniformly distributed with respect to direction. The sentence-level representation - as average of the encoder output - suffers from the same issues \cite{gao2019representation}. Thus, the sentence-level representation space is semantically non-smoothing and poorly defined in some areas. This will lead to the phenomenon that some negative samples are not similar to the speech samples, but the calculated cosine similarity is relatively significant, which will further affect the calculation of contrastive loss \cite{cao2022exploring}. Inspired by research in NLP \cite{huang2021whiteningbert,su2021whitening}, we adopt a simple whitening strategy to make the sentence-level representations anisotropic (Figure \ref{fig:iso}b).

The representations for contrastive loss calculation are $\{ (s_x^{(n)},s_z^{(n)})\} _{n = 1}^N$, where $s_x^{(i)} = {\rm{AveragePooling}}(h_x^{(i)})$, $s_z^{(i)} = {\rm{Whitening}}({\rm{AveragePooling}}(h_z^{(i)}))$, $s_x^{(i)} \in 1 \times d,s_z^{(i)} \in 1 \times d$. The text representation is fixed, while the speech representation is dynamic. So, we only treat the text representations $s_z^{(k),k \ne i}$ as negative samples to ensure the consistency of negative sample representation. Moreover, we set up a First-In-First-Out (FIFO) queue \cite{he2020momentum} to store the text representations of the previous batch to decouple the relationship between the number of negative samples and the batch size. The contrastive loss for coarse granularity is as follows: 

\begin{equation}
\begin{split}
&{{{\cal L}}_{coarse}} = \\
&{ - \frac{1}{N}\sum\limits_{i = 1}^N {\log \frac{{{e^{sim(s_x^i \cdot s_z^i/\tau )}}}}{{\sum\nolimits_{j = 1}^N {{e^{sim(s_x^i \cdot s_z^j/\tau )}}}  + \sum\nolimits_{k = 1}^K {{e^{sim(s_x^i \cdot s_z^k/\tau )}}} }}}} 
\end{split}
\end{equation}                         
where $sim(\cdot)$ is the cosine similarity function, $N$ is the batch size, $\tau $ is the temperature and $K$ is the number of negative samples in the queue. 

\noindent\textbf{Whitening} Given a set of sentence representations $\{ {q_i}\} _{i = 1}^N,{q_i} = {\rm{AveragePooling}}(h_{}^{(i)})$, a linear transformation is performed to ensure the mean value is zero and the covariance is the identity matrix.    

\begin{equation}
{{\tilde q_i} = ({q_i} - \mu )\bm{W}}
\end{equation}
where $\bm{W}$ is the transform matrix, and $\mu $ is defined as the mean vector of the entire embedding set $\{ {q_i}\} _{i = 1}^N$, i.e., $\mu  = \frac{1}{N}\sum\limits_{i = 1}^N {{q_i}} $. We denote the original covariance matrix of $\{ {q_i}\} _{i = 1}^N$ as $\textstyle{\bm{\sum}}$.

The transformed covariance matrix $\bm{ \sum \limits^ \sim}$ of $\{ {\tilde q_i}\} _{i = 1}^N$ need to be the identity matrix:

\begin{equation}
\textstyle{\bm{ \sum \limits^ \sim}} = \bm{W^T} \textstyle{\bm{\sum}} \bm{W} = \bm{I}
\end{equation}

Therefore,
\begin{align}
\textstyle{\bm{\sum}}   &= \bm{({W^T})^{ - 1}}\bm{W} \notag \\
  &= \bm{({W^{ - 1}})^T}\bm{W^{ - 1}}
\end{align}
where $\bm{ \sum}$ is a positive definite symmetric matrix. SVD decomposition is calculated to get the solution:
\begin{align}
\textstyle{\bm{\sum}}  &= \bm{U\Lambda {U^T}} \notag \\
\bm{W} &= \bm{U\sqrt {{\Lambda ^{ - 1}}}}
\end{align} 
   
\subsection{Fine Granularity Contrastive Learning}
\label{sec:Fine Granularity Contrastive Learning}
ST is a generative task, which means each speech frame in the encoder should have precise semantics. Nevertheless, better overall representation does not guarantee more accurate fine granularity representation. We need more advanced processing. The length inconsistency of speech and text makes fine granularity contrast less straightforward than coarse granularity. The alignment between speech frames and text tokens must be obtained. We propose a maximum similarity method to find this correspondence in an unsupervised manner to avoid force alignment, as the alignment model is not always available.

We get the cosine similarity matrix $\Delta  \in {^{{T_x} \times {T_z}}}$ based on the encoded representations $\{ (h_x^{(n)},h_z^{(n)})\} _{n = 1}^N$. ${T_x},{T_z}$ are the length of speech frames and text tokens, respectively. Each speech frame is matched to the text token with the maximum similarity. We denote the \emph{j}-th frame representation of the \emph{i}-th speech as $h_{x,j}^{(i)}$. This process can be described as follows:

\begin{equation}
pos_{x,j}^{(i)} = \{ h_{z,m}^{(i)}|\mathop {\arg \max \limits_m sim(h_{x,j}^{(i)} \cdot h_{z,m}^{(i)})}, m = 1,2, \cdots {T_z}\} 
\end{equation} 
where $h_{x,j}^{(i)}$ is the \emph{j}-th frame representation of the \emph{i}-th speech, $h_{z,m}^{(i)}$ is the \emph{m}-th token representation of the corresponding transcription. The entire matching process can be quickly calculated through the matrix with a small amount of calculation. Notes that we do not use whitening here to ensure computational instability because the length of the token sequence is limited. Given the correspondence between the speech frame and text token, we can quickly get the fine granularity contrastive loss:

\begin{equation}
\begin{split}
&{{{\cal L}}_{fine}} = \\
&  {- \frac{1}{N}\sum\limits_{i = 1}^N {\sum\limits_{j = 1}^{{T_x}} {\log \frac{{{e^{sim(h_{x,j}^{(i)} \cdot pos_{x,j}^{(i)}/\tau )}}}}{{{e^{sim(h_{x,j}^{(i)} \cdot pos_{x,j}^{(i)}/\tau )}} + \sum\nolimits_{k = 1}^{{K}} {{e^{sim(h_{x,j}^{(i)} \cdot s_z^k/\tau )}}} }}} }}
\end{split}
\end{equation}     
where ${K}$ is the number of negative samples, ${T_x}$ is the length of the frame-level speech and $N$ is the batch size. We use the same negative sample strategy as we use for coarse contrast, because it's conceptually and computationally simpler.    

\subsection{Dropdim}
\label{sec:Dropdim}
Currently commonly used data augmentation strategies in speech include Speed Perturb \cite{ko2015audio} and SpecAugment \cite{park2019specaugment}. However, when using Wav2vec 2.0 as the acoustic feature extractor, the effect of these data augmentation strategies is usually limited. In this paper, we adopt dropdim \cite{zhang2022dropdim}, a recently proposed structured dropout method, as a data augmentation strategy. The key idea is to broke the excessive co-adapting between different embedding dimensions and force the self-attention to encode meaningful features with a certain number of embedding dimensions erased.

\subsection{Training}
The contrastive learning is conducted on the encoder. We combine the word-level knowledge distillation \cite{liu2019end} in the decoder to give multi-level guidance for the training of the ST model. The overall loss is the weighted sum of all previous losses:
\begin{equation}
{{\cal L}} = (1-\gamma ) {{{\cal L}}_{ST}} + \alpha {{{\cal L}}_{coarse}} + \beta {{{\cal L}}_{fine}} + \gamma {{{\cal L}}_{KD}}
\end{equation}
where $\alpha, \beta ,\gamma,$ are hyper-parameter to adjust the weight of each loss. ${{\cal L}}_{KD}$ represents the word-level knowledge distillation loss.

\section{Experiment}

\begin{table}[ht]
\centering
\caption{Statistics of all datasets}
\begin{tabular}{l|cc|cc}
\hline
\multicolumn{1}{c|}{\multirow{2}{*}{Language (EN-)}} & \multicolumn{2}{c|}{MuST-C} & \multicolumn{2}{c}{External MT} \\
\multicolumn{1}{c|}{}                                & hours        & \#sent       & Source          & \#sent        \\ \hline
Germany (DE)                                         & 408h         & 234K         & WMT16             & 4.6M          \\
French (FR)                                          & 492h         & 280K         & WMT14             & 40.8M         \\
Russian (RU)                                         & 489h         & 270K         & WMT16             & 2.5M          \\
Spanish (ES)                                         & 504h         & 270K         & WMT13             & 15.2M         \\
Italian (IT)                                         & 465h         & 258K         & OPUS100         & 1.0M          \\
Romanian (RO)                                        & 432h         & 240K         & WMT16             & 0.6M          \\
Portuguese (PT)                                      & 385h         & 244K         & OPUS100         & 1.0M          \\
Dutch (NL)                                           & 442h         & 253K         & OPUS100         & 1.0M          \\ \hline
\end{tabular}
\label{table:Statistics}
\end{table}

\subsection{Dataset and Processing}
\noindent\textbf{MuST-C} MuST-C \cite{di2019must} is a multilingual dataset extracted from TED talks, including source audio, transcriptions, and text translations. Its source language is English, and the target language cover eight language direction: German (De), French (Fr), Russian (Ru), Spanish (Es), Italian (It), Romanian (Ro), Portuguese (Pt), and Dutch (NL). It is the most extensive training data for speech translation. We select the model according to its performance on the validation set and test it on the tst-COMMON set.

\noindent\textbf{External MT Datasets} The MT model is trained separately and has the same structure as the ST model. It allows us to use parallel sentence pairs in the external MT datasets in addition to the transcription-translation pairs in the ST corpus. Table~\ref{table:Statistics} lists the statistics of all the datasets included.
 
\noindent\textbf{Processing} We use the original 16-bit 16kHz mono-channel audio waveform as speech input. We tokenize and true case all texts via Moses\footnote{https://www.statmt.org/moses/}. In each language direction, we apply BPE \cite{sennrich2015neural} on the combination of source and target text to obtain shared sub-word units with a vocabulary size of 8K.

\subsection{Experimental setups}
\noindent\textbf{Model Configuration} We use Wav2vec 2.0 following the large configuration in \cite{baevski2020wav2vec}, which is self-supervised pretrained on Librispeech \cite{7178964} audio data only\footnote{https://huggingface.co/facebook/wav2vec2-large-960h}. We use Transformer \cite{vaswani2017attention} as the backbone of the model, including 6 encoder layers and 6 decoder layers. We train both small and medium size models. For the small size model, each layer comprises 256 hidden units, 4 attention heads, and 2048 feed-forward hidden units. For the medium size model, the above parameters are set to 512, 8, 2048. The queue size $K$ and temperature $\tau$ are set to 1000 and 0.12, respectively, according to our pilot study. 

\noindent\textbf{MT model Pretrain} The previous work \cite{xu2021stacked} shows a domain mismatch between extra datasets and MuST-C. Therefore, the MT model is first trained on the extra MT datasets and then finetuned on transcription-translation pairs of MuST-C.

\noindent\textbf{E2E-ST Training and Inference} Following the previous work, we initialize the ST model encoder and decoder with the pretrained MT model. During training, we use the Adam \cite{kingma2014adam} optimizer with ${\beta _1} = 0.9,{\beta _2} = 0.98$ and adopt the default learning schedule in ESPnet \cite{inaguma_etal_2020_espnet}. The dropout rate and the value of label smoothing are all set to 0.1. Regarding dropdim described in Section \ref{sec:Dropdim}, we adopt the random mask strategy described in their paper with a mask rate of 0.05. An early stop strategy is adopted during training with three epochs 
patiences. We set $\alpha$, $\beta$ and $\gamma$ to 1.0, 1.0, 0.6 respectively. All models are trained on 2 Nvidia Tesla-V100 GPUs. It takes about one day to converge.

During inference, we average the best 5 checkpoints for evaluation. We use beam search with a beam size of 10, and the length penalty is 0.6. We report the case-sensitive SacreBLEU\footnote{sacreBLEU signature: nrefs:1 | bs:1000 | seed:12345 |
case:mixed | eff:no | tok:13a | smooth:exp | version:2.0.0} \cite{post-2018-call} for fair comparison with previous work. 

\noindent\textbf{Baseline Systems} To verify the effectiveness of our method, we compare with the following E2E-ST systems: STAST \cite{liu2020bridging}, AFS \cite{zhang2020adaptive}, SATE \cite{xu2021stacked}, Dual-Decoder \cite{le2020dual}, XSTNet \cite{ye2021end}, TDA \cite{du2021regularizing}, STEMM \cite{fang2022stemm}, JT-S-MT \cite{tang2021improving}, Chimera \cite{han2021learning}, STPT \cite{tang2022unified}, SpeechUT \cite{zhang2022speechut} and ConST \cite{yan2021consert}. We implement FCCL\_base which has the same model architecture as our proposed FCCL. The only difference is that it is trained without fine and coarse granularity contrastive learning. 
     
\subsection{Main Results}

\begin{table*}[]
\centering
\caption{BLEU scores on MuST-C tst-COMMON set. “External Data” indicates whether the method uses additional data. The superscripts $s$ and $m$ represent the small model and medium model, respectively.}
\label{tab:e2e}
\begin{tabular}{lccccccccccc}
\hline
\multicolumn{1}{l|}{\multirow{2}{*}{Model}} & \multicolumn{2}{c|}{External Data} & \multicolumn{9}{c}{BLEU}                                                                                   \\
\multicolumn{1}{l|}{}                       & Speech  & \multicolumn{1}{c|}{MT}  & EN-DE & EN-FR & EN-RU & EN-ES & EN-IT & EN-RO & EN-PT                  & \multicolumn{1}{c|}{EN-NL} & Avg. \\ \hline
\multicolumn{12}{c}{\emph{w/o external MT data}}                                                                                                                                            \\ \hline
\multicolumn{1}{l|}{STAST \cite{liu2020bridging}}                    & $\times$     & \multicolumn{1}{c|}{$\times$}   & 23.1  & -  & -  & -  & -  & -  & -                   & \multicolumn{1}{c|}{-}  & - \\
\multicolumn{1}{l|}{AFS \cite{zhang2020adaptive}}                    & $\times$     & \multicolumn{1}{c|}{$\times$}   & 22.4  & 31.6  & 14.7  & 26.9  & 23.0  & 21.0  & 30.0                   & \multicolumn{1}{c|}{24.9}  & 23.9 \\
\multicolumn{1}{l|}{SATE \cite{xu2021stacked}}                   & $\times$     & \multicolumn{1}{c|}{$\times$}    & 25.2  & -     & -     & -     & -     & -     & -                      & \multicolumn{1}{c|}{-}     & -    \\
\multicolumn{1}{l|}{Dual-Decoder \cite{le2020dual}}           & $\times$     & \multicolumn{1}{c|}{$\times$}    & 23.6  & 33.5  & 15.2  & 28.1  & 24.2  & 22.9  & 30.0                   & \multicolumn{1}{c|}{27.6}  & 25.7 \\
\multicolumn{1}{l|}{STPT \cite{tang2022unified}}           & $\checkmark$     & \multicolumn{1}{c|}{$\times$}    & -  & \textbf{39.7}  & -  & \textbf{33.1}  &-   & -  & -                   & \multicolumn{1}{c|}{-}  & - \\

\multicolumn{1}{l|}{XSTNet \cite{ye2021end}}                 & \checkmark        & \multicolumn{1}{c|}{$\times$}    & 25.5  & 36.0  & 16.9  & 29.6  & 25.5  & \textbf{25.1}  & 31.3                   & \multicolumn{1}{c|}{30.0}  & 27.5 \\
\multicolumn{1}{l|}{TDA \cite{du2021regularizing}}                    & $\times$        & \multicolumn{1}{c|}{$\times$}    & 25.4  & 36.1  & 16.4  & 29.6  & 25.1  & 23.9  & 31.1                   & \multicolumn{1}{c|}{29.6}  & 27.2 \\
\multicolumn{1}{l|}{STEMM \cite{fang2022stemm}}                  & \checkmark        & \multicolumn{1}{c|}{$\times$}    & 25.6  & 36.1  & 17.1  & 30.3  & 25.6  & 24.3  & 31.0                   & \multicolumn{1}{c|}{30.1}  & 27.5 \\

\multicolumn{1}{l|}{ConST \cite{yan2021consert}}                  & \checkmark        & \multicolumn{1}{c|}{$\times$}    & 25.7  & 36.8  & 17.3  & 30.4  & 26.3  & 24.8  & \textbf{32.0}                  & \multicolumn{1}{c|}{\textbf{30.6}}  & 28.0 \\

\multicolumn{1}{l|}{FCCL$^s$\_base}                  & \checkmark        & \multicolumn{1}{c|}{$\times$}    & {24.9}  & {35.7}  & {17.1}  & {29.9}  & {25.2}  & 23.7  & {30.5}                   & \multicolumn{1}{c|}{{29.6}}  & {27.1} \\ 

\multicolumn{1}{l|}{FCCL$^s$}                  & \checkmark        & \multicolumn{1}{c|}{$\times$}    & {25.7}  & {36.5}  & {17.5}  & {30.4}  & {26.0}  & 24.6  & {31.4}                   & \multicolumn{1}{c|}{{30.3}}  & {27.8} \\ 

\multicolumn{1}{l|}{FCCL$^m$}                  & \checkmark        & \multicolumn{1}{c|}{$\times$}    & \textbf{25.9}  & {36.8}  & \textbf{17.6}  & {30.7}  & \textbf{26.4}  & 25.0  & {31.8}                   & \multicolumn{1}{c|}{{30.5}}  & \textbf{28.1} \\ \hline

\multicolumn{12}{c}{\emph{w/ external MT data}}                                                                                                                                             \\ \hline
\multicolumn{1}{l|}{SATE \cite{xu2021stacked}}                   & \checkmark        & \multicolumn{1}{c|}{\checkmark}    & 28.1  & -     & -     & -     & -     & -     & -                      & \multicolumn{1}{c|}{-}     & -    \\
\multicolumn{1}{l|}{JT-S-MT \cite{tang2021improving}}                & $\times$        & \multicolumn{1}{c|}{\checkmark}    & 26.8  & 37.4  & -     & 31.0  & -     & -     & -                      & \multicolumn{1}{c|}{-}     & -     \\

\multicolumn{1}{l|}{SpeechUT \cite{zhang2022speechut}}                & $\times$        & \multicolumn{1}{c|}{\checkmark}    & \textbf{30.1} & \textbf{41.4}  & -     & \textbf{33.6}  & -     & -     & -                      & \multicolumn{1}{c|}{-}     & -     \\

\multicolumn{1}{l|}{XSTNet \cite{ye2021end}}                 & \checkmark        & \multicolumn{1}{c|}{\checkmark}    & 27.8  & {38.0}  & 18.5  & 30.8  & 26.4  & 25.7  & \textbf{32.4}                   & \multicolumn{1}{c|}{\textbf{31.2}}  & 28.8 \\
\multicolumn{1}{l|}{Chimera \cite{han2021learning}}                & \checkmark        & \multicolumn{1}{c|}{\checkmark}    & 26.3  & 35.6  & 17.4  & 30.6  & 25.0  & 24.0  & 30.2                   & \multicolumn{1}{c|}{29.2}  & 27.3 \\
\multicolumn{1}{l|}{TDA \cite{du2021regularizing}}                    & \checkmark        & \multicolumn{1}{c|}{\checkmark}    & 27.1  & 37.4  & -     & -     & -     & -     & -                      & \multicolumn{1}{c|}{-}     & -    \\
\multicolumn{1}{l|}{STEMM \cite{fang2022stemm}}                  & \checkmark        & \multicolumn{1}{c|}{\checkmark}    & 28.7  & 37.4  & 17.8  & 31.0  & 25.8  & 24.5  & 31.7                   & \multicolumn{1}{c|}{30.5}  & 28.4 \\

\multicolumn{1}{l|}{ConST \cite{yan2021consert}}                  & \checkmark        & \multicolumn{1}{c|}{\checkmark}    & 28.3  & {38.3} & 18.9  & {32.0}  & 27.2  & 25.6  & \textbf{33.1}                   & \multicolumn{1}{c|}{\textbf{31.7}}  & 29.4 \\

\multicolumn{1}{l|}{FCCL$^s$\_base}                  & \checkmark        & \multicolumn{1}{c|}{\checkmark}    & {27.6}  & 36.8 & {18.0}  & {30.3}  & {25.7}  & {25.1}  & 31.0 & \multicolumn{1}{c|}{30.2}  & {28.0} \\ 

\multicolumn{1}{l|}{FCCL$^s$}                  & \checkmark        & \multicolumn{1}{c|}{\checkmark}    & {28.7}  & 37.5  & {19.1}  & {31.2}  & {26.5}  & {26.0}  & 32.1                   & \multicolumn{1}{c|}{31.0}  & {29.0} \\ 

\multicolumn{1}{l|}{FCCL$^m$}                  & \checkmark        & \multicolumn{1}{c|}{\checkmark}    & {29.0}  & {38.3}  & \textbf{19.7}  & {31.9}  & \textbf{27.3}  & \textbf{26.8}  & 32.7 & \multicolumn{1}{c|}{31.6}  & \textbf{29.6} \\ \hline

\end{tabular}
\end{table*}

\noindent\textbf{Comparison with E2E Baselines} Table~\ref{tab:e2e} shows the results on the MuST-C dataset. The previous works have shown that the additional datasets can significantly improve the ST model performance. Therefore, to be fair, we mainly compare prior results both with and without additional MT datasets. (a) Without external MT datasets. FCCL$^s$ obtain an average improvement of 0.7 BLEU compared with FCCL$^s$\_base. This show that contrastive learning can effectively guide the learning of the ST model, leading to better translation performance. When including the results from previous work, FCCL$^m$ outperforms the previous models and achieves new state-of-the-art. Different from STEMM \cite{fang2022stemm}, which adopts the manifold mixup to alleviate the representation discrepancy in an implicit manner, we bridge the modality gap through explicit knowledge transfer with the help of contrastive learning. Thus, FCCL achieves better performance, and the improvement over STEMM is also remarkable. (b) With external MT datasets. Compared to itself, FCCL$^m$ can achieve 1.2/1.5 BLEU improvement in different model sizes when additional MT datasets are available. This demonstrates FCCL ability to leverage additional datasets. Chimera \cite{han2021learning} designed a shared semantic memory through contrastive learning to learn the semantic information shared between modalities. However, it limits the feature output lengths of the two modalities to be consistent, which will sacrifice the MT model performance \cite{yan2021consert}. Our proposed method does not have this limitation. ConST \cite{yan2021consert} introduces a similar idea to bridge the modality gap. Nevertheless, they ignore the nature of ST task and only conduct contrastive learning at sentence-level. In contrast, we propose to give more clear guidance at both sentence- and frame-level, which achieves better performance. Our model is worse than SpeechUT \cite{zhang2022speechut} and STPT \cite{tang2022unified}. However, they mainly focus on the pre-training procedure. Our proposed method is orthogonal with theirs, and we will investigate how to combine them together in the future. 

\begin{table}[]
\centering
\caption{Comparison with cascaded models on MuST-C En-De and En-Fr tst-COMMON set.}
\label{tab:cas}
\begin{tabular}{cc|cc}
\hline
\multicolumn{2}{c|}{\multirow{2}{*}{Model}}            & \multicolumn{2}{c}{BLEU} \\
\multicolumn{2}{c|}{}                                  & En-De       & En-Er      \\ \hline
\multicolumn{1}{c|}{\multirow{3}{*}{Cascaded}} & XSTNet \cite{ye2021end}    & 25.2        & 34.9       \\
\multicolumn{1}{c|}{}                          & STEMM \cite{fang2022stemm} & 27.5        & -          \\
\multicolumn{1}{c|}{}                          & SATE \cite{xu2021stacked}  & 28.2        & -          \\ \hline
\multicolumn{1}{r|}{End-to-end}                & FCCL$^m$ & \textbf{29.0}        & \textbf{38.3}       \\ \hline
\end{tabular}
\end{table}

\noindent\textbf{Comparison with Cascaded Baselines} To further validate the effectiveness of our proposed method, we compare with several strong cascaded baseline systems, all of which are trained with additional datasets. As described in Table~\ref{tab:cas}, our proposed method outperform the cascade model and achieve better performance.

\section{Analysis}
\subsection{How to get the MT model?}
When transferring knowledge from the MT to the ST model, an important question is how to get the MT model. In this section, we conduct some analysis of this. Pretrain represents we pretrain the MT model and freeze its parameters, which is used in this paper. Joint represents that we jointly train the MT and ST task in a shared model. Baseline means the ST model is trained only with the cross-entropy loss.

As shown in Table~\ref{tab:mt}, the performance of the joint-trained model on the ST task is better than baseline, but worse than pretrain. Although joint training can use the MT task as an additional task to constrain the parameter solution space of the ST task, the opposite is not always true. In the absence of sophisticated techniques, the performance of the joint-trained model on the MT task is often inferior to the MT model trained alone. Although a tuned semantic space benefits the early learning of ST, the performance decreasing of the joint-trained model on the MT task in later training will have a more significant negative impact. In the future, we will study how to combine the advantages of pretrain and joint training.

\begin{table}[]
\centering
\caption{Model performance under different training methods.}
\label{tab:mt}
\setlength{\tabcolsep}{4.0mm}{
\begin{tabular}{c|clc}
\hline
\multirow{2}{*}{Models} & \multicolumn{3}{c}{Task} \\ \cline{2-4} 
                        & ST       &     & MT      \\ \hline
Baseline                & 23.79    &     & 28.10   \\
joint                   & 24.43    &     & 27.47   \\
pretrain                & 25.71    &     & 28.10   \\ \hline
\end{tabular}}
\end{table}

\subsection{Is it necessary to initialize the ST model with a pretrained MT model?}
In this paper, we propose to improve the ST model performance by explicit knowledge transfer. However, the ST encoder and decoder are initialized from a pretrained MT model. Initialization corresponds to implicit knowledge transfer. In this section, we study the effect of initialization on FCCL.  

\begin{figure}[!t]
\centering
\includegraphics[scale=0.6]{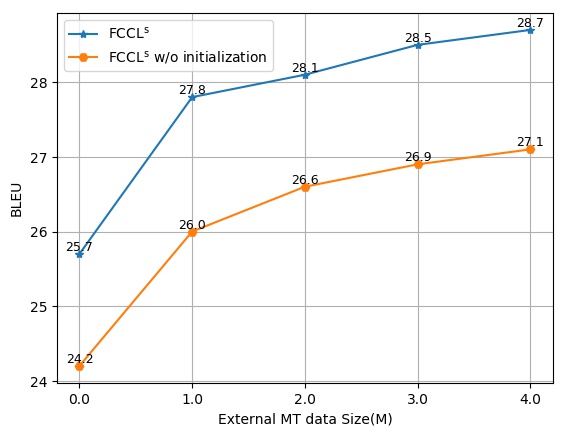}
\caption{BLEU scores on MuST-C En-De tst-COMMON against the size of external MT data.}
\label{fig:initia}
\end{figure}

As shown in Figure~\ref{fig:initia}, initialization can yield improvements over 1.0 BLEU. This can be attributed to the fact that initialization from a pretrained MT model can eliminate the randomness caused by different starting points of optimization \cite{rofin2022linear}, making the parameter space of the ST model to be simpler than that of the model trained from scratch. Although some studies in NLP show that initialization from a pretrained model (such as BART \cite{lewis2020bart} and mBART \cite{liu2020multilingual}) will undermine the downstream task performance under high-resource settings (data pairs $>=$ 10M) \cite{ding2022improving}, its condition is too rigorous to observe a similar phenomenon in the ST task. On the one hand, as shown in Table~\ref{table:Statistics}, the amount of data in the MuST-C dataset is limited. On the other hand, the external supervision signal used to guide the ST model learning may still not be strong enough. That results in FCCL not being completely free of the effects of initialization. We will conduct further research in the future, making FCCL fully “explicit”.

\subsection{How does the whitening method work?}
\label{sec:whitening}
In this paper, we use the whitening method to alleviate the representation degeneration of the MT model. A natural question is how it works and affects the calculation of contrastive loss. We analyzed the cosine similarity distribution to make some explanations. Specifically, we randomly select a speech sample from the MuST-C En-De tst-COMMON set as the anchor sample, and 1000 transcription samples. For each sample, we average its encoder output over the time dimension to get the overall representation. The cosine similarity between 1000 samples and anchor sample (speech sample) is $[s_x^1 \cdot s_z^1,s_x^1 \cdot s_z^2, \cdots ,s_x^1 \cdot s_z^j],j = 1, \cdots 1000$, where $s_x^1$ denotes the speech representation, $s_z^j$ is \emph{j}-th text representation. $s_z^1$ is the corresponding positive sample of $s_x^1$, and the rest are treated as negative. Then we normalize the cosine similarity $[s_x^1 \cdot s_z^1,s_x^1 \cdot s_z^2, \cdots ,s_x^1 \cdot s_z^j]/s_x^1 \cdot s_z^1,j = 1, \cdots 1000$. 

\begin{figure}[!t]
\centering
\includegraphics[scale=0.6]{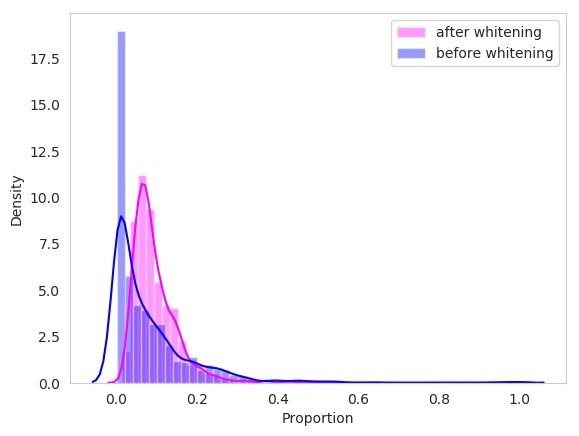}
\caption{Histogram and probability density function (pdf) of the cosine similarity.}
\label{fig:whitening}
\end{figure}

The histogram and probability density function (pdf) of the cosine similarity is shown in Figure~\ref{fig:whitening}. Before whitening, the pdf is a long-tailed distribution, which means that there exist many samples with high resemblance to the anchor sample except the positive sample. Unfortunately, many of these samples are not really similar to the anchor sample. The anisotropy of text representation leads to the appearance of spurious correlation. These long-tailed samples can be eliminated after whitening, indicating that the whitening operation can alleviate the long-tailed problem of cosine similarity distribution. The contrastive loss is based on cosine similarity. Thus, the whitening method can make contrastive learning more precise. 

\subsection{Effectiveness of Each Learning Objective}
In order to analyze the effectiveness of different modules in our method, we conduct an ablation study on the MuST-C En-De dataset. As shown in Table~\ref{tab:ab}, each part of FCCL is necessary and has a positive effect. (1) Removing the fine and coarse granularity contrastive loss brings about a decrease of 0.45 and 0.68 BLEU, respectively, indicating that the contrastive loss can guide the learning of the ST model. (2) Moreover, removing both of them will cause further performance degradation (0.8 BLEU), indicating that fine and coarse contrastive learning are complementary. Without coarse granularity contrast, fine granularity contrast produces slight effect on model performance, because good features will not be learned if incorrect correspondence is extracted. (3) When the contrast loss is calculated directly without whitening, 0.69 BLEU reduction was observed. This valid the analysis in Section~\ref{sec:whitening}, showing that whitening can alleviate the long-tailed problem of cosine similarity distribution and make contrastive learning more precise. (4) When knowledge distillation is removed, the model performance drops by 0.67 BLEU. In FCCL, knowledge distillation and contrastive learning guide the ST model from the decoder and encoder outputs, respectively. They are complementary, and removing either one will adversely affect the model. (5) It is worth noting that when dropdim is removed, the model performance has a significant drop, about 0.95 BLEU. This can be attributed to the fact that dropdim in FCCL can not only enhance the generalization ability of the model, but also generate more challenging representations to enhance the effect of contrastive learning.

\begin{table}[]
\centering
\caption{BLEU scores on MuST-C En-De tst-COMMON set when different parts are removed.}
\setlength{\tabcolsep}{5.0mm}{
\label{tab:ab}
\begin{tabular}{l|l}
\hline
Model    & BLEU  \\ \hline
FCCL$^s$    & 25.71 \\
\quad\quad -fine contra.  & 25.26(-0.45)  \\
\quad\quad -coarse contra.    & 25.03(-0.68) \\
\quad\quad\quad -fine contra.    & 24.91(-0.80) \\
\quad\quad -whitening      & 25.02(-0.69) \\
\quad\quad -knowledge distillation      & 25.04(-0.67) \\
\quad\quad -dropdim & 24.76(-0.95) \\ \hline
\end{tabular}}
\end{table}

\subsection{Model Hallucinations}
\label{sec:Model Hallucinations}

\begin{figure}[!t]
\centering
\includegraphics[scale=0.55]{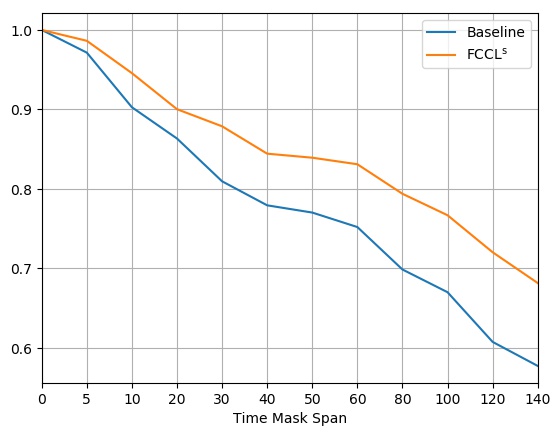}
\caption{Model performance on MuST-C En-De tst-COMMON under different strength perturbations. The horizontal coordinate represents the time mask span. The vertical coordinate represents the ratio between the BLEU value under the perturbed input and the BLEU value under the unperturbed input.}

\label{fig:hall}
\end{figure}

\begin{figure*}[!t] 
  \centering 
  \subfloat{ 
    \label{fig:subfig:a} %% label for first subfigure 
    \includegraphics[width=2.3in]{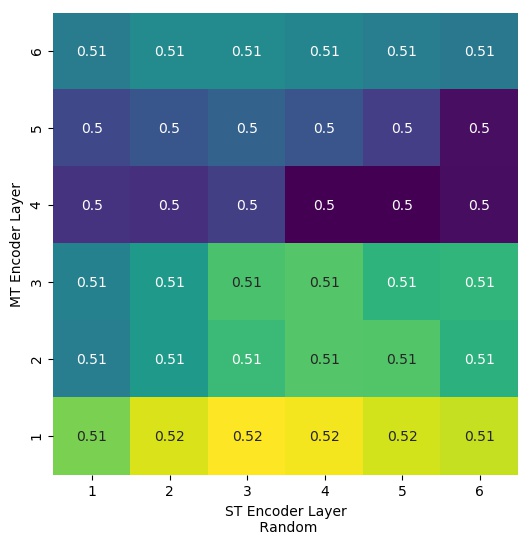}
 	}
  \subfloat{ 
    \label{fig:subfig:b} %% label for second subfigure 
    \includegraphics[width=2.3in]{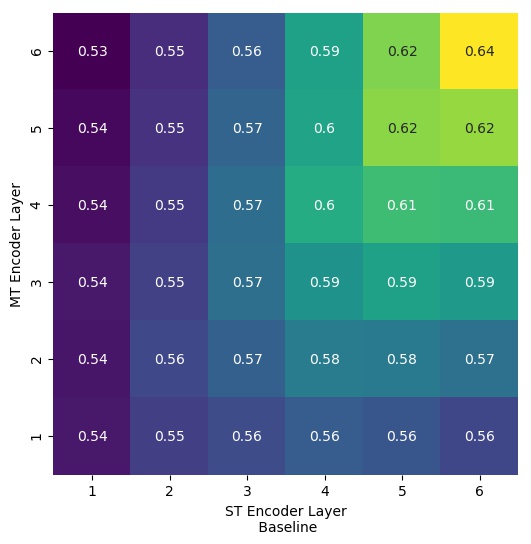}} 
  \subfloat{ 
    \label{fig:subfig:c} %% label for second subfigure 
    \includegraphics[width=2.52in]{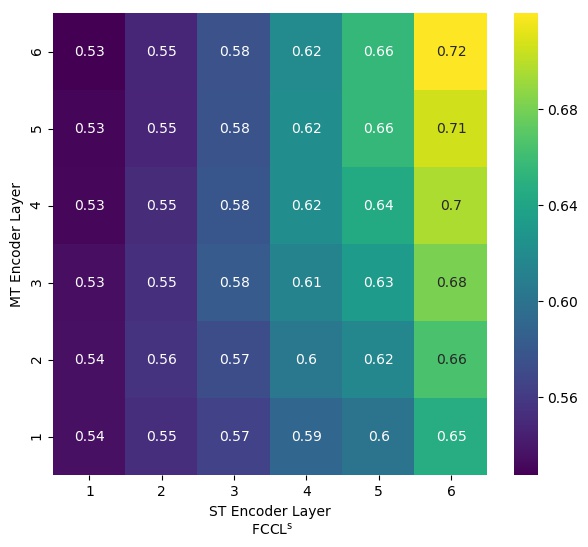}} 
  \caption{PWCCA between encoder states of different layers of the ST and the MT model.} 
  \label{fig:cca} %% label for entire figure 
\end{figure*}

The attention mechanism of a model might not reflect a model actual inner reasoning. In MT, Lee \cite{katherine2019hallu} proposed the concept of hallucinations. A model hallucinates if a small perturbation in the input causes a sharp change in the output, indicating that the model does not really pay attention to the input. He selects the most common words from the corpus as perturbations. However, giving the exact definition in speech requires force alignment, which is costly. Instead, a simple perturbation method is considered in this paper. We adopt the time mask in SpecAugment \cite{park2019specaugment} as a perturbation method to test the model performance under perturbation.

As shown in Figure \ref{fig:hall}, the blue and yellow curves represent the performance of the Baseline and FCCL$^s$ under different strength perturbations compared to the unperturbed input. The baseline represents an ST model trained only with the end-to-end cross-entropy loss ${{{\cal L}}_{ST}}$. The larger the time mask span, the more noise in the input. It can be seen that the performance of FCCL$^s$ decreases more slowly when the perturbation strength gradually increases, suggesting it is more resilient against perturbations and more attentive to the content of its input.
 
\subsection{Canonical Correlation Analysis}
\label{sec:Canonical}
To further analyze FCCL, we turn to canonical correlation analysis (CCA), which finds a linear transformation that maximizes the correlation between two high-dimensional representations. Raghu \cite{raghu2017svcca} defined each neuron activation vector as its response over a finite set of inputs, and the amount of data determines the dimension of the activation vector. For a given dataset $X = \{ {x_1}, \cdots ,{x_m}\}$ , the activation vector of the neuron $i$ in the layer $l$ is defined as $z_i^l$ i.e., $z_i^l = (z_i^l({x_1}), \cdots ,z_i^l({x_m}))$.           

In CCA, enough data is essential. On the one hand, it can reduce the occurrence of spurious correlation, and on the other hand, it can also ensure the stability of the calculation. We mix the MuST-C En-De tst-COMMON and dev sets to increase the amount of data used to compute CCA. Additionally, we normalize the activation vector values to [-1, 1] as suggested by \cite{kambhatla2022cipherdaug}. For each input, we average the representation over the time dimension as the neuron response to the corresponding input. We save the activations for each encoder layer for three types ST models, namely Baseline, FCCL$^s$, and Random models. The parameters of random model are entirely random. We use it as a comparison to exclude the influence of spurious correlations caused by insufficient data. The baseline model is the same as defined in Section~\ref{sec:Model Hallucinations}. We also save activations for the MT model. Then we compute the projection weighted CCA (PWCCA) \cite{morcos2018insights} between activations from those layers. A high value indicates that the representations between layers are linearly related, implying that the layers capture similar information.

Figure \ref{fig:cca} plots the PWCCA between different layers of different networks, and we can observe three exciting phenomena.

 (1) For the Random model, it already has certain similarities with the MT model. On the one hand, the features extracted by Wav2vec 2.0 already contain rich acoustic information. On the other hand, the limited amount of data used to calculate PWCCA leads to spurious correlations.

 (2) FCCL forces more layers to learn semantic information. The output of the sixth layer of the MT model encoder contains the high-level semantic information of the text, and the lower layer output contains grammatical structure information. In Baseline, only the last two layers are used to learn semantic information (only the fifth and sixth layers in the Baseline model have the max correlations with the sixth layer of the MT model). In contrast, the last four layers are used to learn semantic information in the FCCL$^s$ (the last four layers in FCCL$^s$ model have the max correlations with the sixth layer of the MT model).

 (3) Compared with the Baseline, FCCL can effectively improve the correlation between each layer and the semantic representation of the text (the first row of FCCL$^s$ has significantly higher values than the Baseline). Since a network capacity is limited, FCCL$^s$ can free up its capacity from learning grammatical structure information and force more layers to learn semantic information. In this way, the model can extract representations containing more semantic information, and its performance also gets improvement. This proves that our method can effectively bridge the gap between the two modalities, validating our model design.

\subsection{Is the maximum similarity method reasonable?}
When conducting fine granularity contrastive learning, we need to get the correspondence between the speech frames and text tokens. To avoid the use of an extra alignment model, we propose a maximum similarity method. Two natural questions are whether this approach makes sense and whether it incurs additional computing costs because it is online. 

\begin{figure}[!t]
\centering
\includegraphics[scale=0.43]{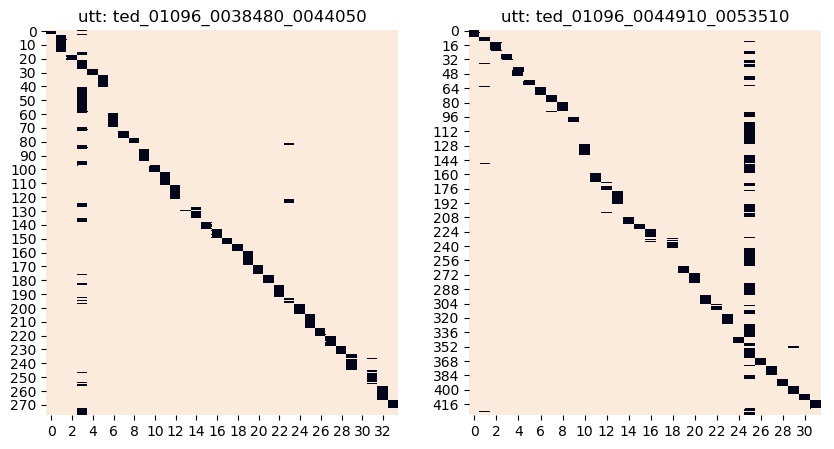}
\caption{Visualization of similarity matrix after masking.}
\label{fig:fine}
\end{figure}

\textbf{We answer the first question by performing fine granularity visualization.} We randomly select paired speech-transcription from MuST-C En-De tst-COMMON set, then calculate the similarity matrix $\Delta  \in {^{{T_x} \times {T_z}}}$ between speech and transcription representation. As shown in Figure~\ref{fig:fine}, the overall correspondence is monotonic. Some non-monotonic alignment can be attributed to the presence of silence and noise frames in speech. The model learns to correspond these frames to unique text token.

\textbf{We answer the second question by computing the FLOPs.} The FLOPs with and without fine-grained contrastive learning are 21.67G and 21.66G, respectively. With an increase of only 0.01G FLOPs, fine granularity contrastive learning improves the performance by 0.45 BLEU (as shown in Table~\ref{tab:ab}). In general, by adopting this method, we can effectively and efficiently find the correspondence between speech frame and text token in an unsupervised manner with negligible latency overhead.

\subsection{Visualization of Coarse Granularity Alignment}
We randomly select 30 speech-transcription pairs from MuST-C En-De tst-COMMON set, and then apply T-SNE \cite{van2008visualizing} to the vector representations of these samples to reduce the dimension to two. Note that these vector representations are obtained by averaging the encoder outputs over the time dimension.

\begin{figure}[!t]
\centering
\includegraphics[scale=0.6]{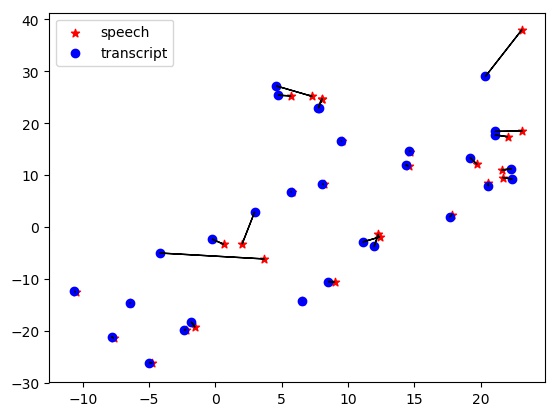}
\caption{Visualization the sentence-level representation.}
\label{fig:coarse}
\end{figure}

The results are visualized in Figure~\ref{fig:coarse}. Each speech-transcription pair is connected by a solid line. It can be intuitively seen from the figure that most paired speech-transcription are projected together, and some even overlap with each other. This proves that FCCL is capable of bridging the representation divergence of the two modalities. In addition, some speech representations in the figure are still clustered together, mainly because we compute the contrastive loss across modalities and not within the modal. Thus, the speech representations do not show good uniformity.

\subsection{Visualization of Model Learning Dynamic}
\begin{figure}[t]
\centering
\includegraphics[scale=0.58]{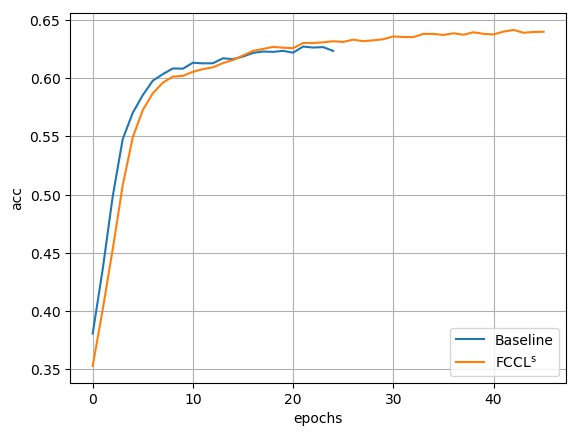}
\caption{Visualization of training process.}
\label{fig:dya}
\end{figure}

Figure~\ref{fig:dya} shows the learning dynamics. The blue and yellow curves represent the accuracy of the Baseline and FCCL$^s$ on the validation set, respectively. Although the FCCL$^s$ achieves better performance, its performance in the early stage is worse than the Baseline. One reason is that we use dropdim to increase the learning difficulty of the model. In addition, the quality of the speech representation in the early stage is poor, so the maximum similarity method cannot find the correct correspondence. A possible solution is to discard the fine granularity contrastive loss in the early stage of training and add it until the ST model has a specific representation ability.

\section{Conclusion and Future Work}
In this paper, we propose a cross-modal multi-grained contrast learning method, FCCL, for \textbf{\emph{explicit knowledge transfer}} from the MT to the ST model. In addition, we propose whitening to solve the representation degeneration of text representation in the MT model. Experiments on the MuST-C dataset in all 8 languages demonstrate the effectiveness of our method. Additional experiment analysis and visualization show that FCCL is capable of bridging the speech-text representation gap and exhibits stronger robustness.

Although our method exhibits the desired effect, it relied on transcription to guide the extract of speech representation during training. For the more than 7000 languages and dialects worldwide, most of them do not have corresponding translations or even transcriptions, and our method does not work in such scenarios. More effective strategies to improve the quality of E2E-ST need to be explored in the future.

\bibliographystyle{IEEEtran}
\bibliography{custom1}

\end{document}